\documentclass[a4paper,11pt]{article}

\usepackage{fourier}
\usepackage{color}
\usepackage{graphicx}
\usepackage{url}
\usepackage[affil-it]{authblk}
\usepackage[T1]{fontenc}
\usepackage{times}
\usepackage{amsmath}
\usepackage{tabularx} 
\usepackage[font={small}]{caption}
\usepackage{tikz}
\usepackage{floatrow}
\newsavebox{\measurebox}
\usepackage{color,soul}
\usepackage{comment}
\usepackage{subcaption}

\newcommand{\textOverlay}[2]{%
  \begin{tikzpicture}
    \node at (1.2cm, -1.3cm) (image) {#1};
    \draw node [black] {#2};
  \end{tikzpicture}%
}

\begin{document}

\title{Automated Identification of Trampoline Skills\\Using Computer Vision Extracted Pose Estimation}

\author{Paul W. Connolly, Guenole C. Silvestre and Chris J. Bleakley}
\affil{School of Computer Science, University College Dublin, Belfield, Dublin 4, Ireland.}
\date{}
\maketitle
\thispagestyle{empty}

\maketitle

\begin{abstract}
A novel method to identify trampoline skills using a single video camera is proposed herein.
Conventional computer vision techniques are used for identification, estimation, and tracking of the gymnast's body in a video recording of the routine.
For each frame, an open source convolutional neural network is used to estimate the pose of the athlete's body.
Body orientation and joint angle estimates are extracted from these pose estimates.
The trajectories of these angle estimates over time are compared with those of labelled reference skills.
A nearest neighbour classifier utilising a mean squared error distance metric is used to identify the skill performed.
A dataset containing 714 skill examples with 20 distinct skills performed by adult male and female gymnasts was recorded and used for evaluation of the system.
The system was found to achieve a skill identification accuracy of 80.7\% for the dataset.
\end{abstract}

\section{Introduction}
Originating in the 1930s, trampolining became a competitive Olympic sport in Sydney 2000.
In competition, athletes perform a routine consisting of a series of skills performed over a number of jumps.
The skills are scored by human judges according to the Trampoline Code of Points \cite{cop17}.
Although more explicit and objective judging criteria have been introduced in recent years, the scores awarded can still vary between judges leading to highly contentious final decisions.
Eliminating human error by means of reliable, automated judging of trampoline routines is desirable.
Herein, we describe a first step towards this goal: a novel automated system for identification of trampoline skills using a single video camera.
Identification of skills is necessary prior to judging since different skills are scored in different ways.

While still a challenging problem, identification of trampoline skills from video has been enabled by recent advances in human pose estimation. In \cite{Andriluka2014}, improved accuracy over model-based approaches was achieved with the introduction of convolutional neural network (ConvNet) based estimation.
Estimators such as this rely on new ConvNet algorithms coupled with recent gains in GPU performance.
In addition, the introduction of larger, more varied general pose datasets \cite{modec13,Johnson11}, leveraging crowd-sourced annotation, has vastly increased the quantity of training data available.

To the best of the authors' knowledge, no previous work has been reported on identification of trampolining skills from video.
The closest previous work on identification of trampoline skills required the gymnast to wear a full-body motion capture suit containing inertial sensors \cite{Helten2011}.
Wearing special suits is cumbersome and is not allowed in competition due to the strict rules regarding gymnast attire \cite{cop17}.
Previous work on automated judging of rhythmic gymnastics from video was reported in \cite{DiazPereira2014}.
However, their method differs from the method used in this work.

The algorithm proposed herein consists of a number of stages.
The bounding box of the gymnast is extracted using conventional image processing techniques. The pose of the athlete is subsequently determined, allowing body orientation and joint angles to be estimated.
The angle trajectories over time are compared with those obtained for reference skills.
The skill performed is identified as the nearest neighbour in the reference dataset based on a mean square error metric.
The system was evaluated using a large number of video recordings capturing the movements of male and female gymnasts performing trampoline routines.
A wide variety of skills, lighting conditions, and backgrounds were recorded.
The gymnasts did not wear special clothes or markers.
The camera was placed side-on to the performance, in the same position as a human judge.

The structure of the paper is as follows.
In section \ref{sec:background}, background information on trampolining is given.
In section \ref{sec:related_work}, further detail is provided on approaches to analysis of sporting movement and pose estimation using video recordings.
The proposed algorithm is described in section \ref{sec:prop_method}.
Section \ref{sec:exp_procedure} discusses the experimental procedure and organisation of the dataset.
The experimental results and discussion are provided in section \ref{sec:results}.
Conclusions, including future work, follow in section \ref{sec:conclusion}.

\section{Background}\label{sec:background}

A trampoline routine consists of a sequence of high, continuous rhythmic rotational jumps
performed without hesitation or intermediate straight bounces. The routine should show good form, execution, height, and maintenance of height in all jumps so as to demonstrate control of the body during the flying phase.
A competition routine consists of 10 such jumps, referred to, in this work, as skills.
For simplicity, a straight bounce is taken to be a skill.
A competitor can perform a variable number of straight bounces before the beginning of a routine (so called in-bounces) while an optional straight bounce (out-bounce) can be taken after completing a routine, to control height before the gymnast is required to stop completely.

\begin{figure}[!htbp]
    \captionof{figure}{Take-off and landing positions: (a) feet, (b) seat, (c) front and (d) back. Trampoline shapes: (e) tuck, (f) pike, (g) straddle and (h) straight.}
    \label{tab:novice}
    \centering
    \begin{tabularx}{\textwidth}{|X|X|X|X|}
        \hline
        \begin{minipage}{.2\textwidth}
            \textOverlay{\includegraphics[trim={0 1cm 0 1cm},clip,width=0.8\textwidth]{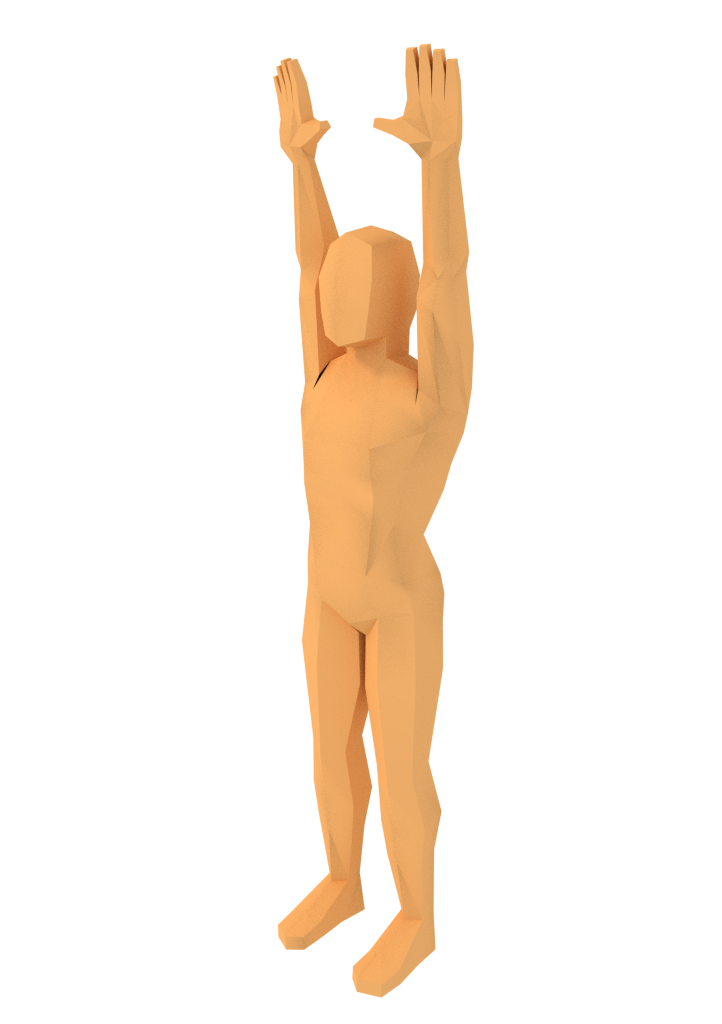}}{(a)}
        \end{minipage}
        &
        \begin{minipage}{.2\textwidth}
            \textOverlay{\includegraphics[trim={0 1cm 0 1cm},clip,width=0.8\textwidth]{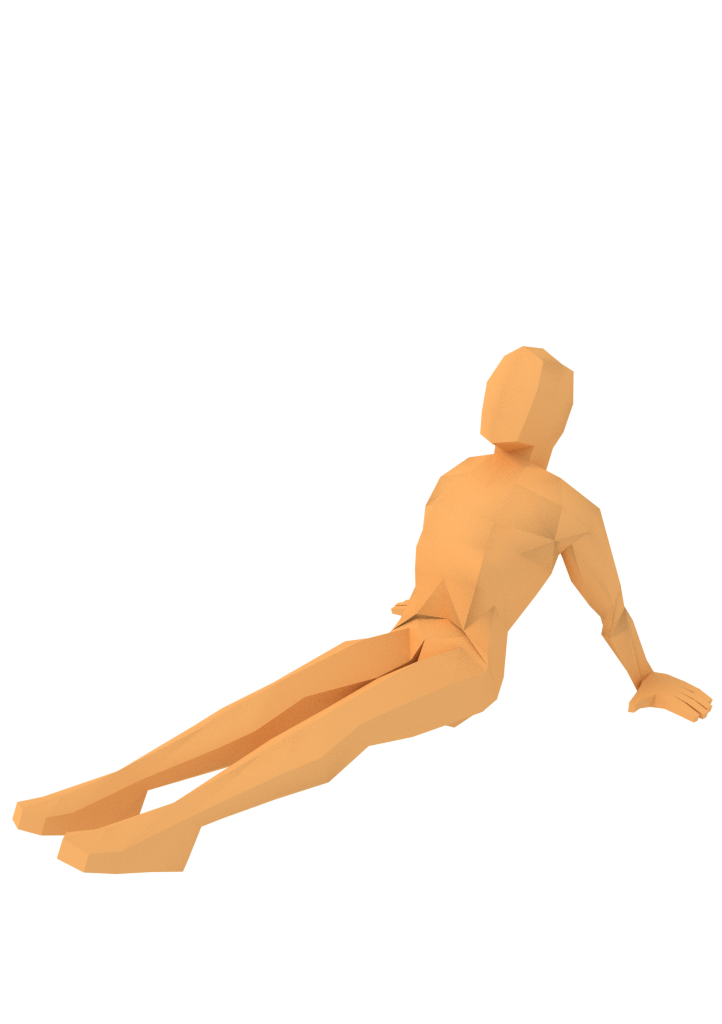}}{(b)}
        \end{minipage}
        &
        \begin{minipage}{.2\textwidth}
            \textOverlay{\includegraphics[trim={0 1cm 0 1cm},clip,width=0.8\textwidth]{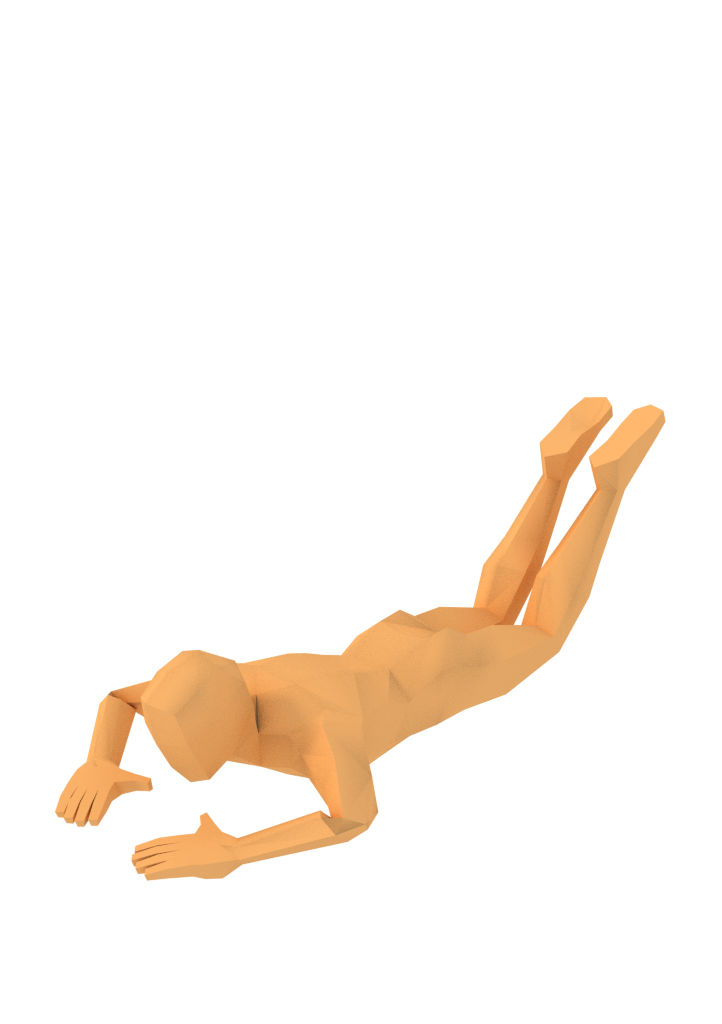}}{(c)}
        \end{minipage}
        &
        \begin{minipage}{.2\textwidth}
            \textOverlay{\includegraphics[trim={0 1cm 0 1cm},clip,width=0.8\textwidth]{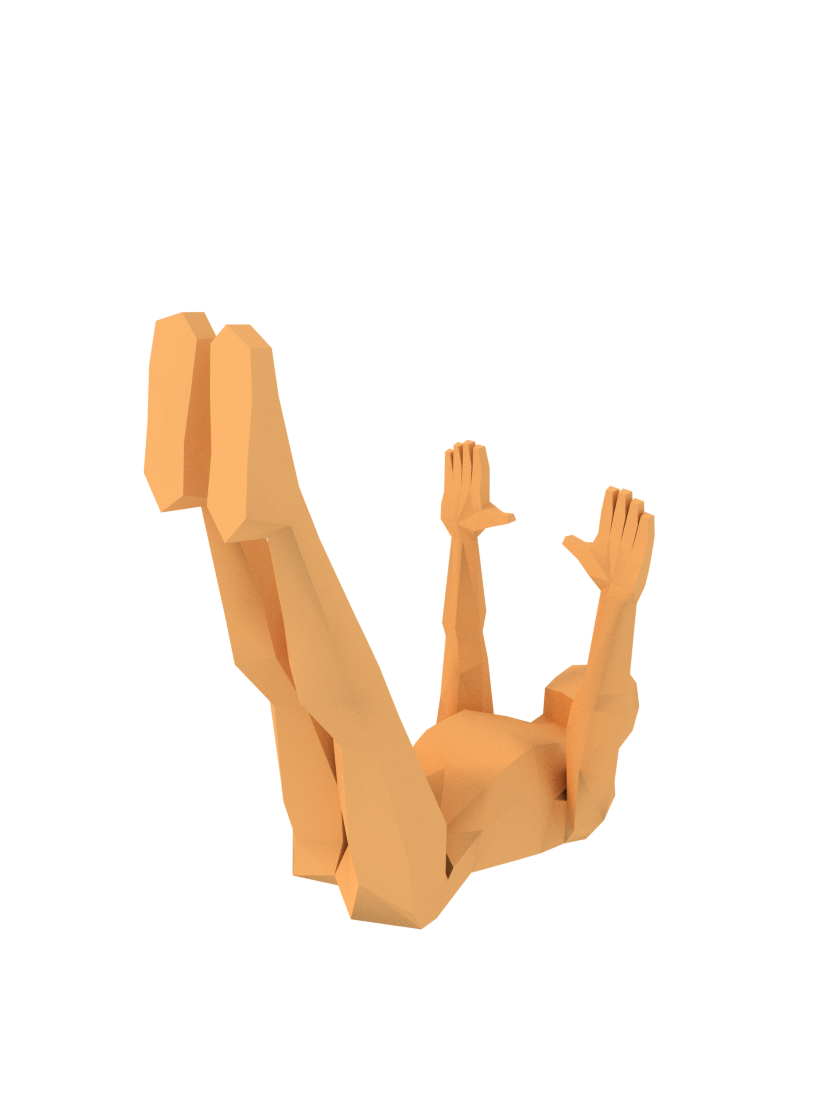}}{(d)}
        \end{minipage}
        \\
        \hline
        \begin{minipage}{.2\textwidth}
            \textOverlay{\includegraphics[trim={0 1cm 0 1cm},clip,width=0.8\textwidth]{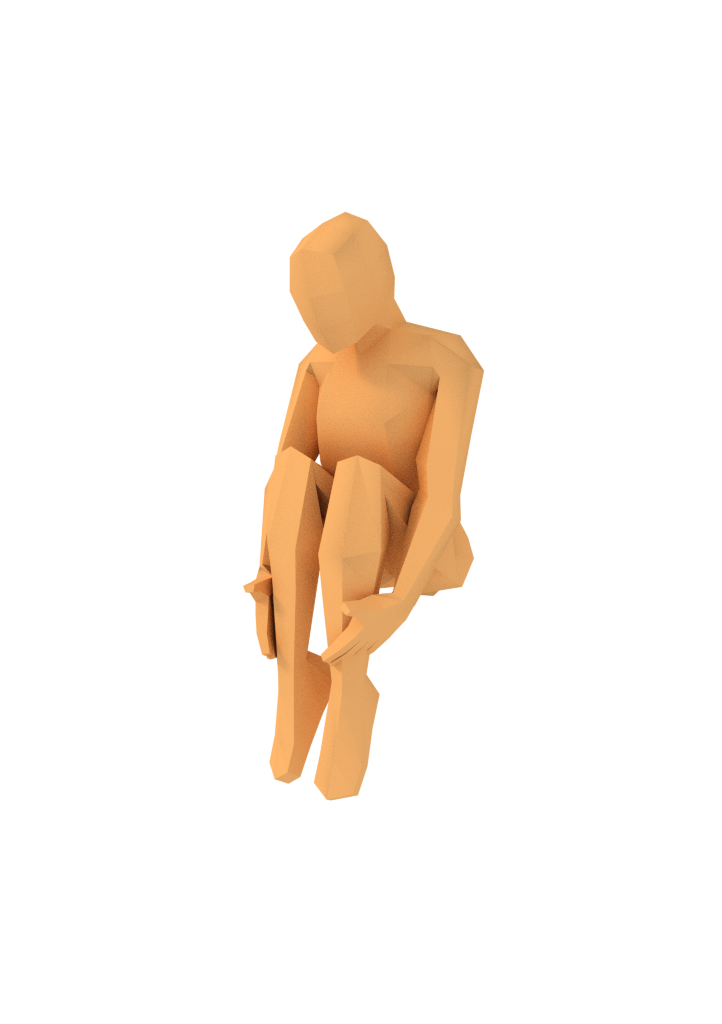}}{(e)}
        \end{minipage}
        &
        \begin{minipage}{.2\textwidth}
            \textOverlay{\includegraphics[trim={0 1cm 0 1cm},clip,width=0.8\textwidth]{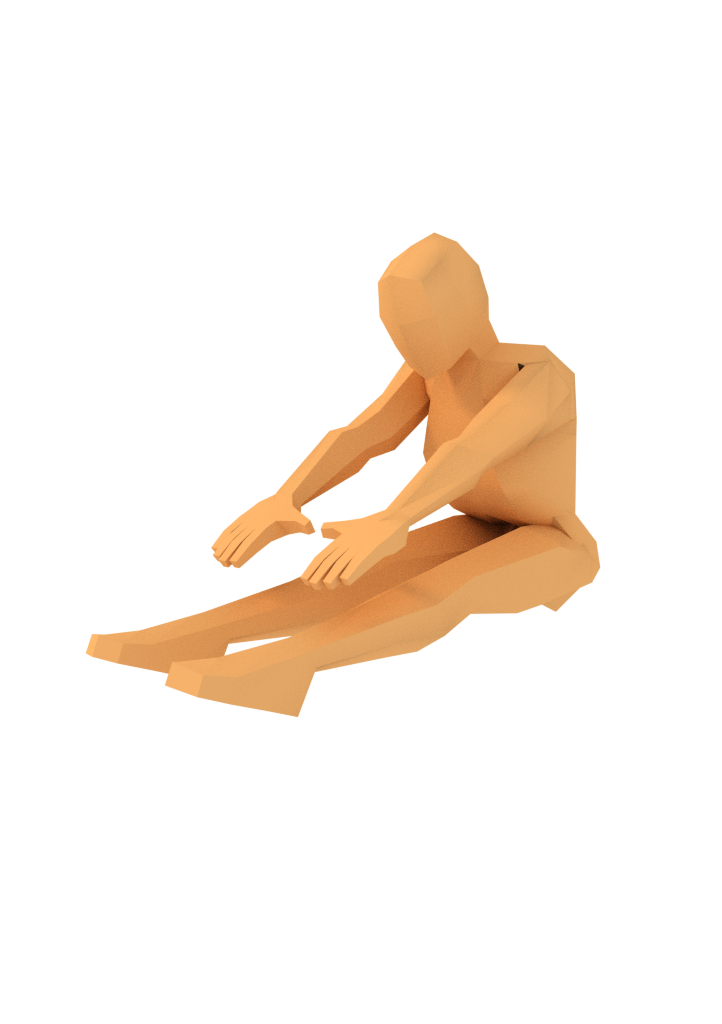}}{(f)}
        \end{minipage}
        &
        \begin{minipage}{.2\textwidth}
            \textOverlay{\includegraphics[trim={0 1cm 0 1cm},clip,width=0.8\textwidth]{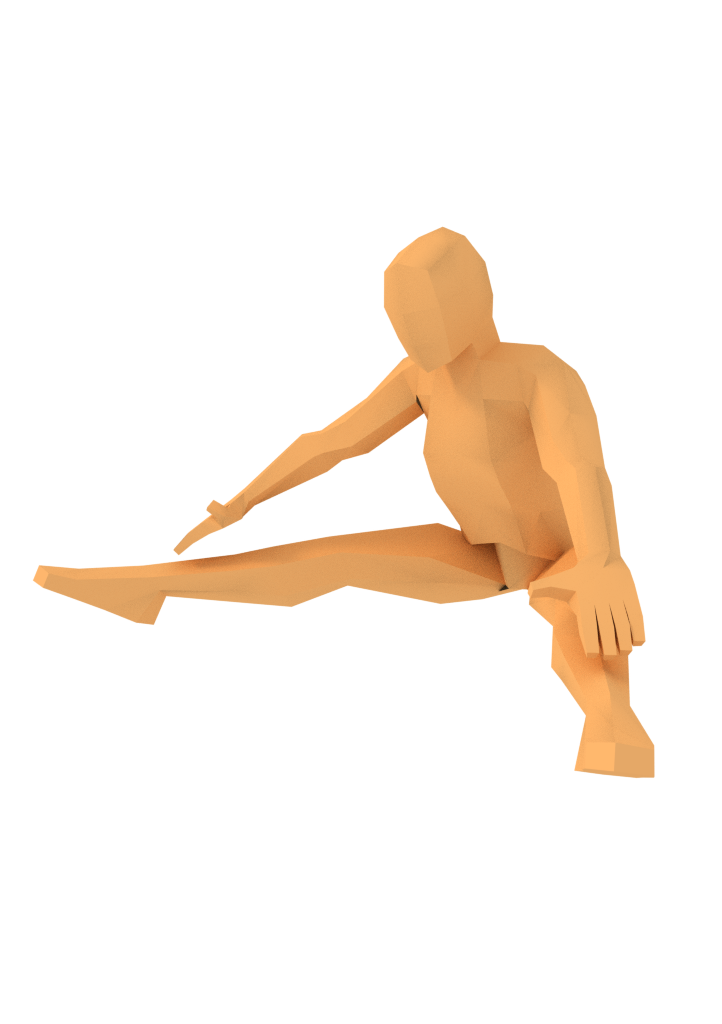}}{(g)}
        \end{minipage}
        &
        \begin{minipage}{.2\textwidth}
            \textOverlay{\includegraphics[trim={0 1cm 0 1cm},clip,width=0.8\textwidth]{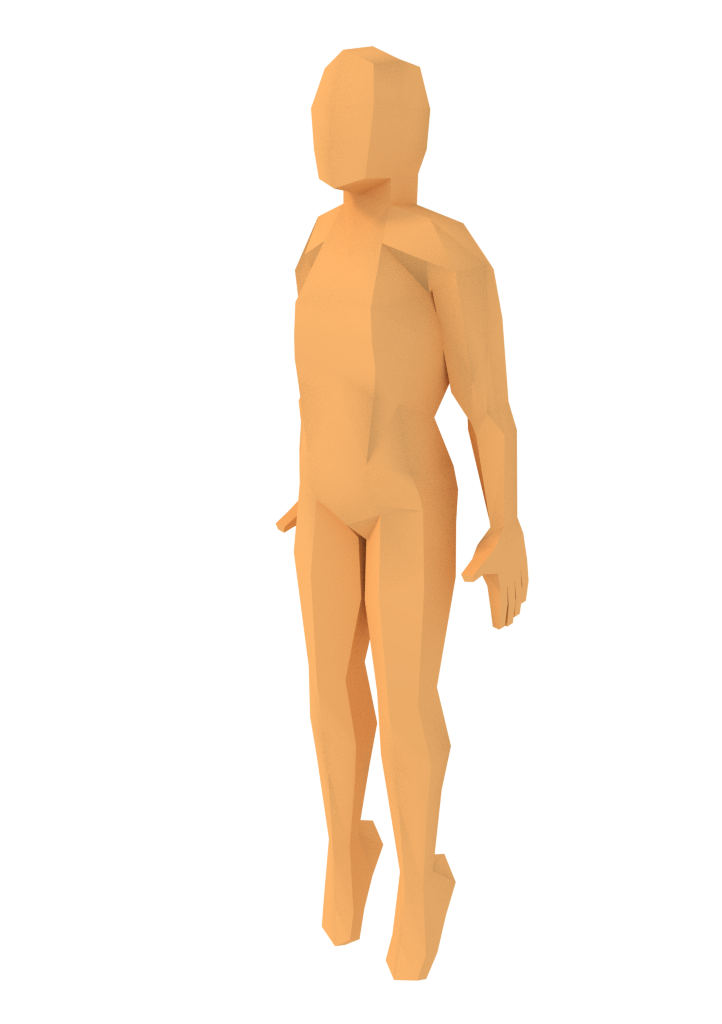}}{(h)}
        \end{minipage}
        \\
        \hline
    \end{tabularx}
    \label{tab:blender}
\end{figure}

Skills involve take-off and landing in one of four positions: feet, seat, front, or back. Rotations about the body's longitudinal and lateral axes are referred to as twist and somersault rotations, respectively. Skills combine these rotations with a body shape: tuck, pike, straddle, or straight. These take-off and landing positions and shapes are illustrated in Figure~\ref{tab:blender}.

The score for a performance is calculated as the sum of four metrics: degree of difficulty (tariff), execution, horizontal displacement, and time of flight.
Degree of difficulty is scored based on the difficulty of the skill performed. For example, a full somersault is awarded more points than a three-quarter somersault. The tariff assigned is found by a simple look-up based on skill identification. Examples of tariff scores can be seen in Table \ref{tab:skills}.
The execution score is allocated based on how well the skill was judged to be performed.
The horizontal displacement and the time of flight are measured electronically using force plates on the legs of the trampoline.

\section{Related Work}\label{sec:related_work}

One of the problems with the capture of trampoline skills is the large performance space. Elite performers can reach up to 10m in height. Tracking such a large volume is prohibitively difficult for many existing motion capture solutions including RGB-D devices such as the Microsoft Kinect.

In \cite{Helten2011}, inertial sensors were used to measure body point acceleration and orientation. The gymnast was required to wear a body suit containing ten inertial measurement units. The sensor data streams were transformed into a feature sequence for classification. For each skill, a motion template was learned. The feature sequence of the unknown trampoline motions were compared with a set of skill templates using a variant of dynamic time warping. The best accuracy achieved was 84.7\% over 14 skill types.

A survey of vision-based methods for general human motion representation, segmentation and recognition can be found in \cite{Weinland2011224}.
In \cite{DiazPereira2014}, judging of rhythmic gymnastics skills from video was investigated. The movement of the gymnast was tracked using optical flow. Velocity field information was extracted across all frames of a skill and projected into a velocity covariance eigenspace. Similar movements were found to trace out unique, but similar, trajectories. New video recordings were classified based on their distance from reference trajectories. The system's specificity was approximately 85\% and the sensitivity was approximately 90\% for the 3 skills considered.



Human pose estimation is the process of estimating the configuration of the body, typically from a single image.
Robust 2D pose estimation has proven to be a powerful starting point for obtaining 3D pose estimates for human bodies.
An overview of the 2D pose estimation problem and proposed methods can be found in \cite{sigal11,Poppe2007}.
Model-based methods have been successful for images in which all the limbs of the subject are visible. However, they are unsuitable for the side-on view of a trampoline routine where self-occlusions are inherent. ConvNet based systems do not assume a particular explicit body model since they learn the mapping between image and body pose. These machine learning based techniques provide greater robustness to variations in clothing and accessories than model-based approaches.
The MPII benchmark \cite{Andriluka2014} has been used to access the accuracy of pose estimators.
The model-based approach described in \cite{PishchulinICCV13} achieved an accuracy of 44.1\% whereas the ConvNet based method proposed in \cite{hourglass} achieved 90.9\%.

The work described herein differs from previous work in that the system performs skill identification for trampolining using a single monocular video camera.
The work takes advantage of recently developed, high accuracy, open source ConvNet based pose estimators.
The Stacked Hourglass Network \cite{hourglass} and MonoCap \cite{monocap} methods were selected for estimation and filtering of the 2D pose, respectively.

In the Stacked Hourglass Network, 2D pose estimates are provided by a ConvNet architecture where features are processed across all scales and consolidated to best capture the spatial relationships of the body parts. Repeated steps of pooling and upsampling, in conjunction with intermediate supervision, out-perform previous methods.
In MonoCap, 3D pose is estimated via an expectation-maximization algorithm over a sequence of images with 2D pose predictions. Conveniently, the 2D joint location uncertainties can be marginalized out during inference.

\section{Proposed Algorithm}
\label{sec:prop_method}

\begin{figure}[t]
    \centering
    \includegraphics[width=0.9\linewidth]{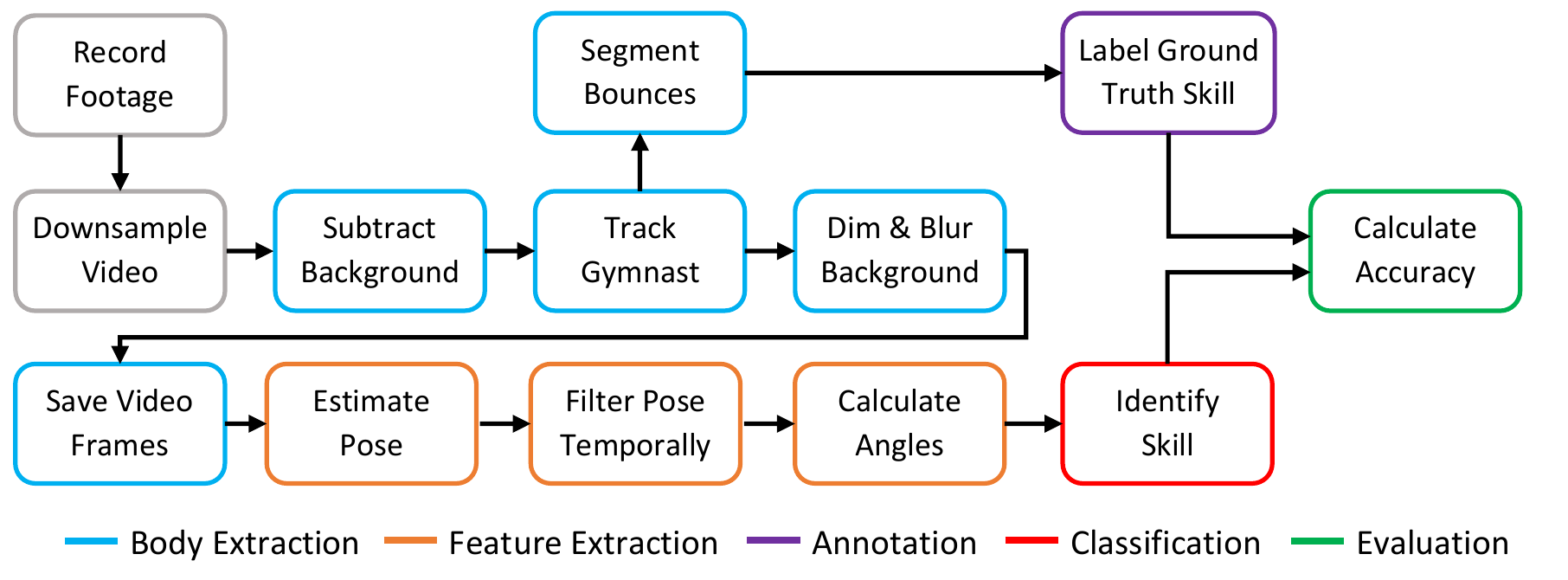}
    \caption{Flow chart illustrating the proposed method.}
    \label{fig:flow_chart}
\end{figure}

The complete algorithm is illustrated in Figure \ref{fig:flow_chart}.
Video is recorded and pre-processed to reduce resolution and remove audio.
After pre-processing, the body extraction stage identifies and tracks the convex hull of the athlete over all video frames.
The video is segmented according to the detected bounces.
The feature extraction stage estimates the pose of the athlete and from this, the body orientation and joint angles in each frame.
Based on the extracted feature angles, classification is performed to identify the skill.
In our experiments, the accuracy of the system was evaluated by comparing the detected skills to manually marked ground truth.
The algorithm stages are explained in more detail in the following sections.

\subsection{Body Extraction}

\begin{figure}[b]
    \centering
    \begin{subfigure}[b]{0.24\textwidth}
        \includegraphics[width=\textwidth]{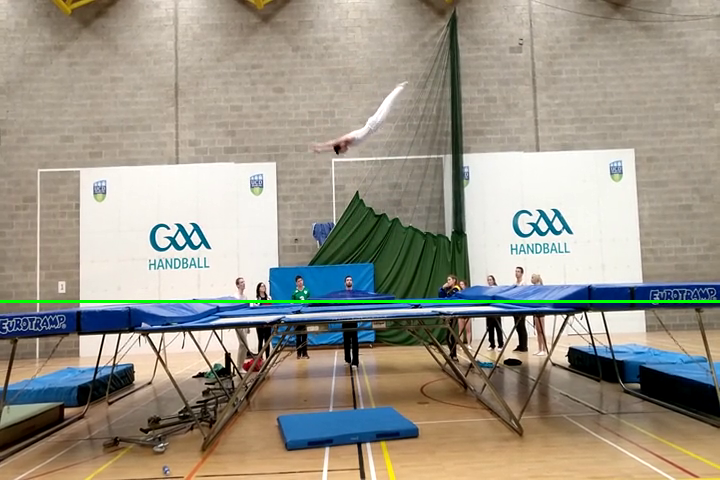}
        \caption{}
    \end{subfigure}
    \hfill
    \begin{subfigure}[b]{0.24\textwidth}
        \includegraphics[width=\textwidth]{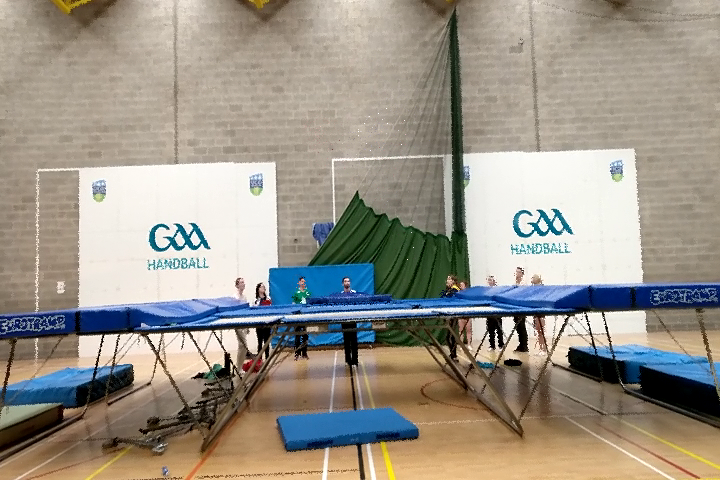}
        \caption{}
    \end{subfigure}
    \hfill
    \begin{subfigure}[b]{0.24\textwidth}
        \includegraphics[width=\textwidth]{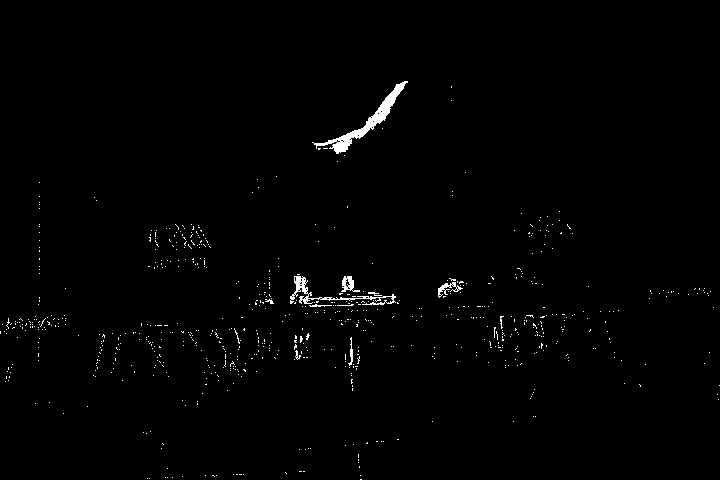}
        \caption{}
    \end{subfigure}
    \hfill
    \begin{subfigure}[b]{0.24\textwidth}
        \includegraphics[width=\textwidth]{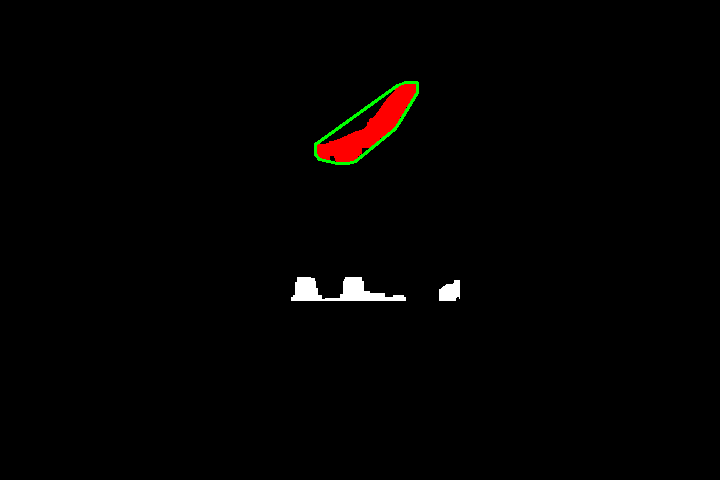}
        \caption{}
    \end{subfigure}
    \caption{Processed images. (a) Original frame. (b) Background model. (c) Foreground mask. (d) Body silhouette and convex hull after erosion and dilation.}
    \label{fig:preprocessing}
\end{figure}

The top of the trampoline is identified based on its hue characteristics and is presented as a best guess on a user interface that allows the position to be fine tuned.
The gymnast is tracked by assuming that they are the largest moving object above the trampoline.
A background subtractor generates a foreground mask for each frame.
All static image components over multiple frames are taken to be part of the background.
The camera is assumed to be static without changing focus during the recording.
The foreground mask is eroded for one iteration and dilated for ten iterations with a $2\times 2$ kernel.
The largest segment of this morphed mask is taken to be the silhouette of the gymnast.
The method of moments is used to determine the centroid of this silhouette.
The video is segmented into individual skills based on the position of this centroid.
A peak detection algorithm identifies the local minima of the vertical position of the centroid.
These local minima are taken to indicate the start and end frames of a skill.
A threshold is applied to peaks between the local minima to identify the start and end jumps of the routine.
The convex hull of the silhouette is used to generate a bounding box for the athlete's image.
The bottom of the bounding box is compared to the position of the top of the trampoline to detect the contact phase of a bounce.
Examples of the application of this method are shown in Figure \ref{fig:preprocessing}.

Images of the body are saved for frames in which the athlete is not in contact with the trampoline.
The maximum size of the bounding box across all frames of the routine is found.
Each image is squarely cropped to this size, centred on the centroid of the gymnast.
Based on the extracted foreground mask, the background of each image is blurred and darkened.
This helps to reduce the number of incorrect pose estimates.

\subsection{Feature Extraction}

\begin{figure}[!ht]
    \centering
    \includegraphics[width=1\linewidth]{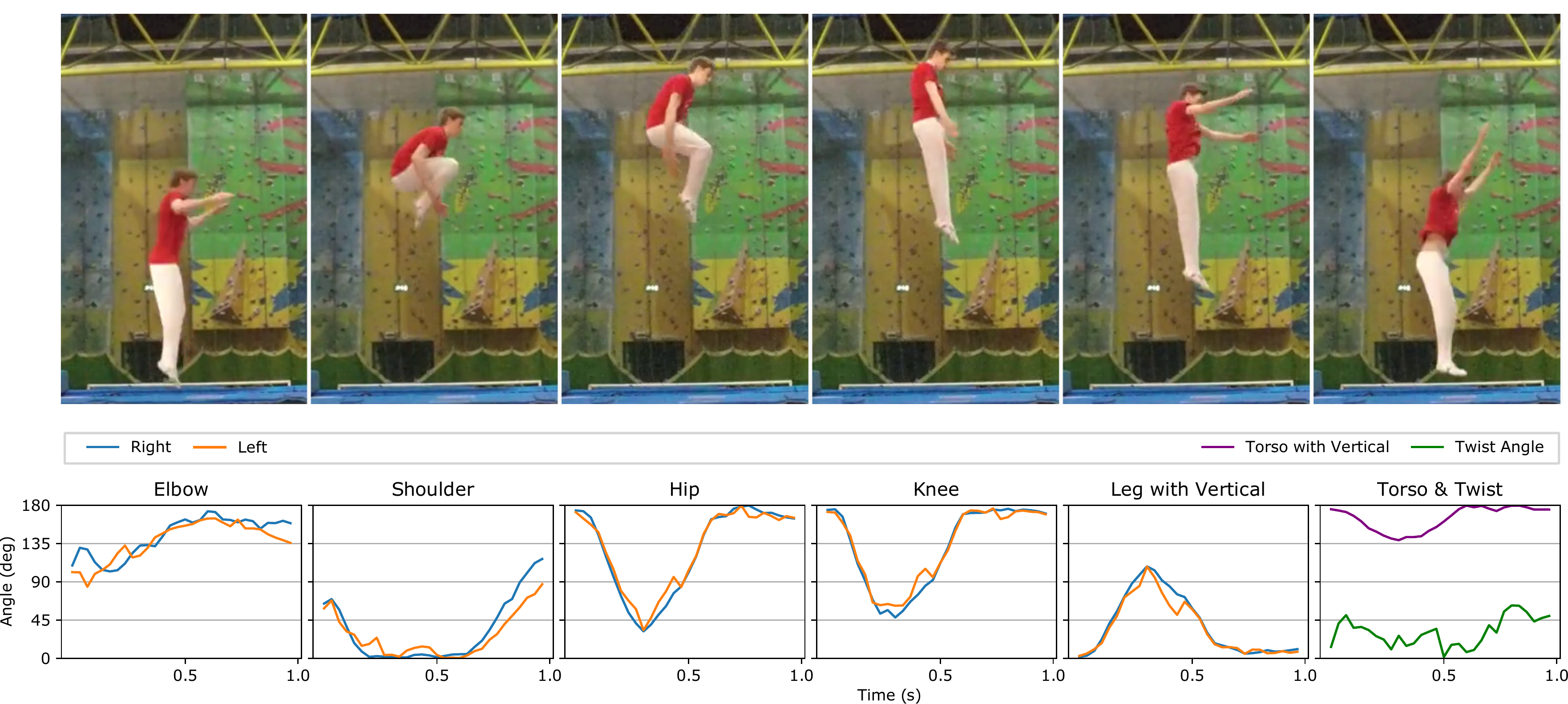}
    \caption{The motion sequence of a tuck jump with the estimated angles shown beneath.}
    \label{fig:skill_filmstrip}
\end{figure}

The Stacked Hourglass Network and MonoCap are used for 2D pose estimation and filtering, respectively.
The 2D pose estimator generates pose predictions for 16 joint locations.
The 3D pose estimator is then used to filter the 2D pose predictions across the sequence of images.
From the smoothed 2D pose, the 2D joint angles and orientation angles that represent the athlete's body position are calculated.
These feature angles are denoted as
$\boldsymbol{\theta}_i$ for $i \in [1 \ldots M]$ where $M = 12$ is the total number of feature angles.
Each of the $M$ feature angles is part of a time series $\theta_i(t)$, where $t$ is the frame number, $t \in [1 \ldots T]$.
The angles are listed in Table \ref{tab:angles_list} and example trajectories can be seen in Figure \ref{fig:skill_filmstrip}.

\begin{table}[t]
    \caption{Feature angles by name and index.}
    \label{tab:angles_list}
    \small
    \begin{tabular}{|c||ccccccc|}
        \hline
            $i$ &
            1, 2&
            3, 4&
            5, 6&
            7, 8&
            9, 10&
            11&
            12\\
        \hline
            $\theta_i(t)$ &
            R, L Elbow &
            R, L Shoulder &
            R, L Hip &
            R, L Knee &
            R, L Leg &
            Torso &
            Twist \\
        \hline
    \end{tabular}
\end{table}

Twist around the body's longitudinal axis is estimated from the 2D distance between the pose points labelled as right and left shoulder. The shoulder separation in the image is at a maximum when the gymnast's back or front is facing the camera and is approximately zero when sideways to the camera. By finding the maximum 2D separation over the whole routine, the separation can be normalised to a value between 0\textdegree and 180\textdegree. In this way, the angle does not depend on the size of the performer.


\subsection{Classification}
The $M$ feature angle trajectories are compared to those in a labelled reference set by calculation of the mean squared error (MSE).
The observed skill is identified as equivalent to the reference giving the minimum MSE.
The feature angle trajectories of the references $\boldsymbol{\theta}^R_i(t)$ are aligned through re-sampling by means of interpolation so as to have the same number of data points $T$ as the observed angle trajectory $\boldsymbol{\theta}^O_i(t)$.


\begin{equation} \displaystyle \label{eq:mse}
    \mbox{MSE} = \frac{1}{TM} \sum_{t=1}^{T} \sum_{i=1}^{M} \left( \boldsymbol{\theta}^O_i(t) - \boldsymbol{\theta}^R_i(t) \right)^2
\end{equation}

\section{Experimental Procedure}
\label{sec:exp_procedure}

\subsection{Data Acquisition}
The procedure for data collection was submitted to and approved by the UCD Office of Research Ethics.
Videos of routines were recorded at training sessions and competitions of the UCD Trampoline Club.
Consent was sought from members of the UCD Trampoline Club prior to recording video for the purposes of the project.
Routines were collected in UCD Sports Centre under normal sports hall lighting conditions.
The background was not modified, typically consisting of a brick wall or nets.

The routines were recorded at a resolution of $1920 \times 1080$ at 30 frames per second (fps) using a consumer grade camera with a shutter speed of 1/120 to reduce motion blur.
The camera was positioned at the typical location and viewing angle of the judging panel.
All bounces were within the field of view of the camera.
The video was subsequently downsampled to $896 \times 504$, maintaining a 16:9 aspect ratio, and audio was removed.
These steps significantly reduced data file size and processing time while maintaining usable resolution.
The videos were manually annotated with ground-truth labels by means of a custom built web interface.

\subsection{Datasets}

The resulting dataset consists of 49 routines by 30 adult athletes, 18 male and 31 female, totalling 23 minutes of video.
This contained 28 distinct skills and 771 skill examples.
The names and distribution of these skills are summarised in Table \ref{tab:skills}.

The accuracy of the identification algorithm was tested using 10-fold cross validation.
The skills with fewer than 10 examples were not included in the test, leaving $N = 20$ distinct skills.

In each iteration of the evaluation, a subset of $10$ examples of each skill were randomly selected from the database. Each subset was split evenly to give the number of reference examples $S_r=5$ and the number of test examples $S_t=5$.
The total size of the reference set was $N \times S_r = 100$ skill examples.
The test set was of the same size.
The average accuracy over 20 iterations of the evaluation is reported herein.



\section{Results and Discussion}
\label{sec:results}

The average accuracy of the system was 80.7\% for the 20 distinct skills listed as included in classification in Table \ref{tab:skills}.
The confusion matrix for the experiment is shown in Figure \ref{fig:confusion}.

It was noted that subject identification can sometimes incorrectly focus on people in the background, particularly during seat, front and back landings, when the gymnast becomes obscured by the trampoline bed.
This causes errors in trampoline contact detection resulting in frames without an obvious subject being presented to the pose estimator. The resulting angles are not representative of the skill performed.
This can also cause errors in jump segmentation due to incorrect centroid extraction.
Jump segmentation failed in 2 cases.


Significant confusion in skill identification occurs between FPF (pike jumps shown in Figure \ref{tab:blender}f) and FSF (straddle jumps shown in Figure \ref{tab:blender}g). From a side-on view, it is difficult to distinguish these movements.
Another area of confusion is between the tuck and pike shape of the Barani skill (BRI).
The features which distinguish these shapes are the angles of the hip and knees.
The tuck shape in this skill is often performed loosely.
This results in the angle of the hip being similar to that of the pike shape.
For identification, the angle of the knees becomes the deciding feature and may be overwhelmed by noise from other features.

Use of a support vector machine might improve classification accuracy.
For example, the difficulty in estimating the wrist and ankle joints for the 2D pose estimator can lead to noise in the angles for the elbows and knees.
Weighting these features as less important might improve overall accuracy.

\begin{minipage}[t]{.47\textwidth}
    \begin{table}[H]
        \caption{Skill dataset. ($^\dag$excluded from classification)}
        \label{tab:skills}
        \footnotesize
        \begin{tabular}{lccc}
            \hline
            Skill Name & Code & Tariff & Occurrences\\
            \hline\hline
            Straight Bounce & F0F & 0.0 & 286\\
            Tuck Jump & FTF & 0.0 & 58\\
            Pike Jump & FPF & 0.0 & 40\\
            Straddle Jump & FSF & 0.0 & 42\\
            \hline
            Half Twist Jump & F1F & 0.1 & 18\\
            Full Twist Jump & F2F & 0.2 & 19\\
            \hline
            Seat Drop & F0S & 0.0 & 13\\
            Half Twist to Seat Drop & F1S & 0.1 & 10\\
            Seat Half Twist To Seat & S1S & 0.1 & 24\\
            To Feet from Seat & S0F & 0.0 & 11\\
            Half Twist to Feet from Seat & S1F & 0.1 & 24\\
            \hline
            Front Drop & F0R & 0.1 & 4$^\dag$\\
            To Feet from Front & R0F & 0.1 & 5$^\dag$\\
            Back Drop & F0B & 0.1 & 10\\
            To Feet from Back & B0F & 0.1 & 8$^\dag$\\
            Half Twist to Feet from Back & B1F & 0.2 & 12\\
            \hline
            Front Somersault (Tuck) & FSSt & 0.5 & 4$^\dag$\\
            Front Somersault (Pike) & FSSp & 0.6 & 7$^\dag$\\
            Barani (Tuck) & BRIt & 0.6 & 24\\
            Barani (Pike) & BRIp & 0.6 & 19\\
            Barani (Straight) & BRIs & 0.6 & 9$^\dag$\\
            Crash Dive & CDI & 0.3 & 18\\
            \hline
            Back Somersault (Tuck) & BSSt & 0.5 & 28\\
            Back Somersault (Pike) & BSSp & 0.6 & 18\\
            Back Somersault (Straight) & BSSs & 0.6 & 30\\
            Back Somersault to Seat (Tuck) & BSTt & 0.5 & 10\\
            Lazy Back & LBK & 0.3 & 3$^\dag$\\
            Cody (Tuck) & CDYt & 0.6 & 3$^\dag$\\
            Back Half & BHA & 0.6 & 1$^\dag$\\
            \hline
            Barani Ball Out (Tuck) & BBOt & 0.7 & 7$^\dag$\\
            Rudolph / Rudi & RUI & 0.8 & 3$^\dag$\\
            Full Front & FFR & 0.7 & 1$^\dag$\\
            Full Back & FUB & 0.7 & 2$^\dag$\\
            \hline
        \end{tabular}
    \end{table}
\end{minipage}
\hfill
\begin{minipage}[t]{.42\textwidth}
    \begin{figure}[H]
        \includegraphics[width=\textwidth]{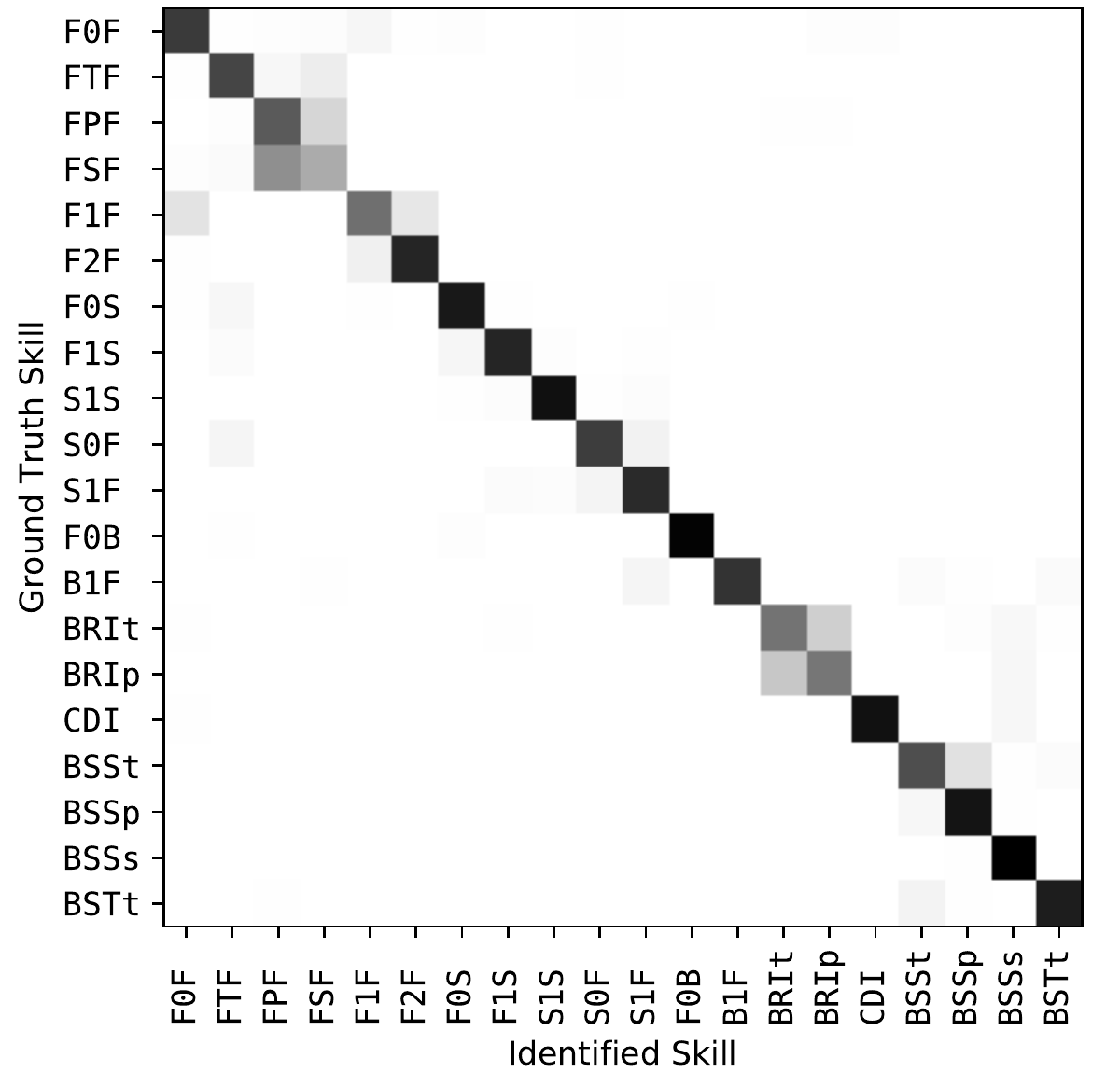}
        \caption{Confusion matrix showing the relative errors for each skill. This is the average of 20 iterations of 10-fold cross validation.}
        \label{fig:confusion}
    \end{figure}
\end{minipage}

\vspace*{1.5ex}

It is likely that accuracy could be improved by increasing the amount of data.
Current pose estimation algorithms take a single image as input.
It seems likely that performance could be improved by tracking pose over a video sequence.
Adding a second video camera pointed towards the front of the gymnast would likely improve accuracy by allowing greater discrimination of motion parallel to the axis between the subject and the first camera.
However, there are issues regarding the extra user effort in setting up the second camera and in synchronisation of the two devices.
Modern trampoline judging systems incorporate force plates for detection of the centrality of landing on the trampoline bed.
Fusing such information with the video data could possibly also result in improved accuracy.

Body extraction was performed at 15 fps on a 2 core Intel i7--3517U 2.4 GHz CPU.
Estimation of pose using the Stacked Hourglass Network ran at 20 fps on an Ubuntu 16.04 with an Nvidia Titan X (Pascal) GPU and a 4 core Intel i7--920 2.67 GHz CPU with default parameter settings.
Execution of the MonoCap algorithm ran at 0.3 fps on the same machine also with default parameters.


\section{Conclusion}
\label{sec:conclusion}
A system for identifying trampolining skills using a single monocular video camera was developed.
The system incorporated algorithms for background subtraction, erosion and dilation, pose estimation, pose filtering and classification.
The system was found to provide 80.7\% accuracy in identifying the 20 distinct skills present in a dataset contain 712 skill examples.

In future work, we plan to extend the classification algorithms to perform automated execution judging.



{\small
\bibliographystyle{apalike}
\bibliography{refs_ieee}
}

\end{document}